\documentclass[runningheads]{llncs}
\usepackage{graphicx}
\usepackage[T1]{fontenc}
\usepackage{graphicx}
\usepackage[table]{xcolor}
\usepackage{amsmath}
\usepackage{amsfonts}
\usepackage{bm}
\usepackage{booktabs}
\usepackage{comment}
\usepackage{multirow}
\usepackage{makecell}
\usepackage{subfig}
\usepackage{afterpage}

\newcommand*{\twoelementtable}[3][l]%
{%
    \begin{tabular}[t]{@{}#1@{}}%
        #2\tabularnewline
        #3%
    \end{tabular}%
}
\newcolumntype{x}{>{\centering\arraybackslash}p{.75cm}}
\newcolumntype{C}[1]{>{\centering\arraybackslash}p{#1}}
\makeatletter
\g@addto@macro{\endtabular}{\rowfont{}}
\makeatother
\newcommand{\rowfonttype}{}
\newcommand{\rowfont}[1]{
   \gdef\rowfonttype{#1}#1%
}

\usepackage{xspace}
\makeatletter
\DeclareRobustCommand\onedot{\futurelet\@let@token\@onedot}
\def\@onedot{\ifx\@let@token.\else.\null\fi\xspace}

\def\eg{\emph{e.g}\onedot} \def\Eg{\emph{E.g}\onedot}
\def\ie{\emph{i.e}\onedot} 
\def\cf{\emph{c.f}\onedot} 
\def\etc{\emph{etc}\onedot} 
\def\wrt{w.r.t\onedot} 

\def\etal{\emph{et al}\onedot}
\makeatother

\begin{document}
\title{Self-supervised 3D Patient Modeling with Multi-modal Attentive Fusion\thanks{Corresponding author: Meng Zheng, email: meng.zheng@uii-ai.com}}
%
%
\author{Meng Zheng\inst{1}\orcidID{0000-0002-6677-2017} \and
Benjamin Planche\inst{1}\orcidID{0000-0002-6110-6437} \and
Xuan Gong\inst{2}\orcidID{0000-0001-8303-633X}
\and
Fan Yang\inst{1}\orcidID{0000-0003-1535-447X}
\and
Terrence Chen\inst{1}
\and Ziyan Wu\inst{1}\orcidID{0000-0002-9774-7770}}
\authorrunning{F. Author et al.}
%
\institute{United Imaging Intelligence, Cambridge MA, USA \and University at Buffalo, Buffalo NY, USA\\
\email{\{first.last\}@uii-ai.com, xuangong@buffalo.edu}}
\maketitle              
\begin{abstract}
3D patient body modeling is critical to the success of automated patient positioning for smart medical scanning and operating rooms. Existing CNN-based end-to-end patient modeling solutions typically require a) customized network designs demanding large amount of relevant training data, covering extensive realistic clinical scenarios (\eg, patient covered by sheets), which leads to suboptimal generalizability in practical deployment, 
b) expensive 3D human model annotations, \ie, requiring huge amount of manual effort, resulting in systems that scale poorly. To address these issues, we propose a generic modularized 3D patient modeling method consists of (a) a multi-modal keypoint detection module with attentive fusion for 2D patient joint localization, to learn complementary cross-modality patient body information, leading to improved keypoint localization robustness and generalizability in a wide variety of imaging (\eg, CT, MRI etc.) and clinical scenarios (\eg, heavy occlusions); and (b) a self-supervised 3D mesh regression module which does not require expensive 3D mesh parameter annotations to train, bringing immediate cost benefits for clinical deployment. We demonstrate the efficacy of the proposed method by extensive patient positioning experiments on both public and  clinical data. Our evaluation results achieve superior patient positioning performance across various imaging modalities in real clinical scenarios.

\end{abstract}

\section{Introduction}
\begin{figure*}[t]
	\centering
	\includegraphics[width=0.9\linewidth]{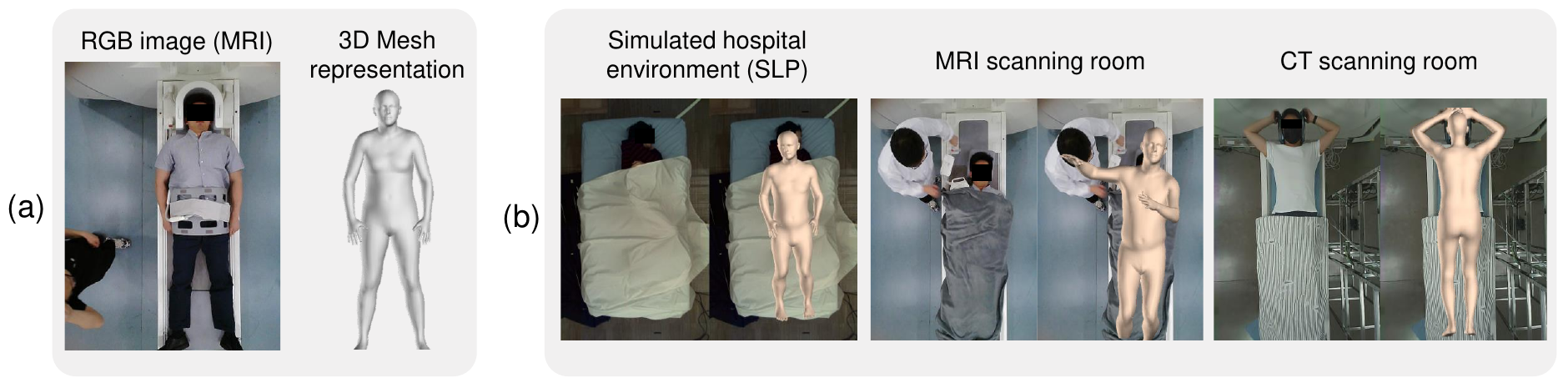}
	\caption{(a) Mesh representation of a patient in MRI scanning room. (b) Failure cases of state-of-the-art mesh regressors (SPIN \cite{kolotouros2019spin}) in challenging clinical scenarios, \eg, simulated hospital environment \cite{SLP_2019}, MRI and CT scanning rooms.}
	\label{fig:fig1}
\end{figure*}

The automatic patient positioning system and algorithm design for intelligent medical scanning/operating rooms has attracted increasing attention in recent years \cite{karanam2020towards,srivastav2018mvor,Kadkhodamohammadi2017_nm,Kadkhodamohammadi2017_tx,avatar_Vivek2017}, 
with the goals of minimizing technician effort, providing superior performance in patient positioning accuracy 
and enabling contactless operation 
to reduce physical interactions and disease contagion between healthcare workers and patients. 
Critical to the design of such a patient positioning system, 
3D patient body modeling in medical environments based on observations from one or a group of optical sensors (\eg, RGB/depth/IR) is typically formulated as a 3D patient body modeling or pose estimation problem \cite{bogo2016keep,kanazawa2018end,kolotouros2019learning} in the computer vision field, defined as follows. 
Given an image captured from an optical sensor installed in the medical environment, we aim to 
automatically estimate the pose and shape---and generate a digital representation---of the patient of interest.
Here we consider 3D mesh representations among several commonly used human representations (\eg, skeleton, contour etc. \cite{survey_2dhpe}), which consist of a collection of vertices, edges and faces and contain rich pose and shape information of the real human body, as demonstrated in Figure \ref{fig:fig1}(a). The 3D mesh estimation of a patient can be found suitable for a wide variety of clinical applications. For instance, in CT scanning procedure, automated isocentering can be achieved by using the patient thickness computed from the estimated 3D mesh \cite{li2007automatic,karanam2020towards}. 
Consequently, there has been much recent work from both algorithm \cite{ching2014patient,kolotouros2019learning,georgakis2020hierarchical} as well as system perspectives \cite{karanam2020towards,Clever_2020_CVPR}.

State-of-the-art patient mesh estimation algorithms \cite{yang2020robust} typically rely on end-to-end customized deep networks, requiring extensive relevant training data for real clinical deployment. For example, training the RDF model proposed in \cite{yang2020robust} requires pairs of multi-modal sensor images and 3D mesh parameters (which are particularly expensive to create \cite{loper2015smpl,loper2014mosh,SMPL-X:2019}).
Moreover, conventional end-to-end 3D mesh estimation methods \cite{kolotouros2019spin,kanazawa2018end,yang2020robust,Multimodal_arxiv2020} assume a perfect person detection as preprocessing step for stable inference, \ie, relying on an efficient person detection algorithm to crop a person rectangle covering the person's full body out of the original image.
Hence, any error during this first person detection step propagates and further impacts the mesh estimation process itself (see Fig. \ref{fig:fig1}(b)), and such detection errors are especially likely to occur when the target patient is \textit{under-the-cover} (\ie, occluded by sheets) or occluded by medical devices. 

\begin{figure*}[t]
	\centering
 	\includegraphics[width=0.8\linewidth]{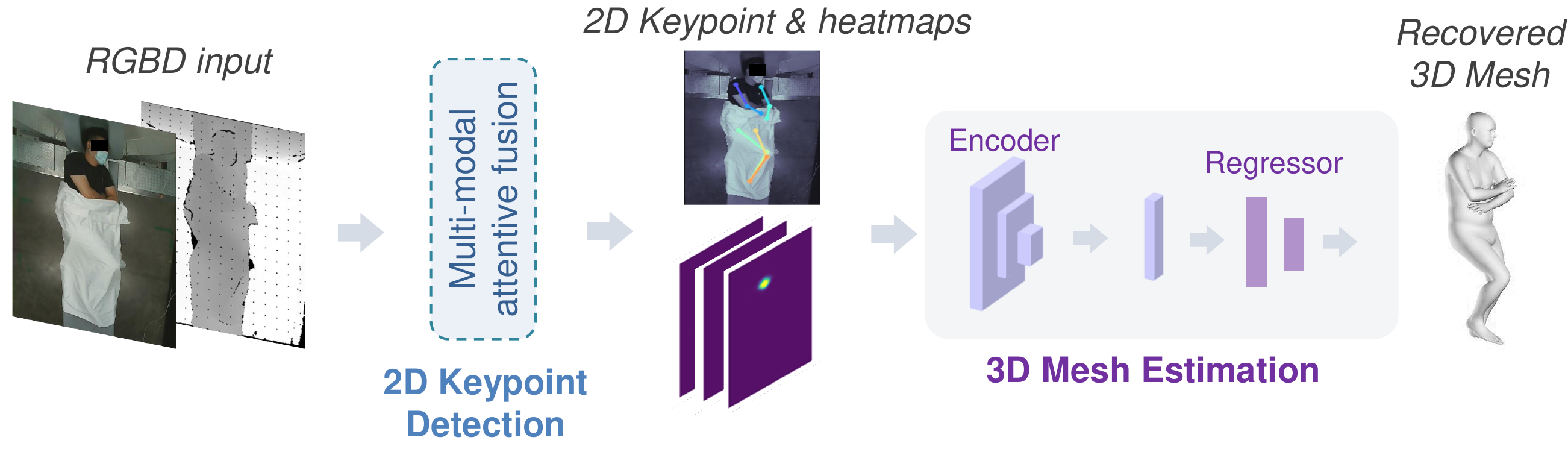}
	\caption{Proposed framework to localize 2D keypoints and infer the 3D mesh.}
	\label{fig:pipeline}
\end{figure*}

We thus propose a multi-modal data processing system that can (a) perform both person detection and mesh estimation, and (b) be trained over inexpensive data annotations.
This system comprises several modules (\cf Figure~\ref{fig:pipeline}). First, we train a multi-modal fused 2D keypoint predictor to learn complementary patient information that may not be available from mono-modal sensors. We then process these 2D keypoints with a novel 3D mesh regressor  
designed to efficiently learn from inexpensively-produced synthetic data pairs in a self-supervised way.
Besides technical contributions within each module, \eg, cross-modal attention fusion for improved joint localization and self-supervised mesh regression (\cf Section \ref{sec:method}), we demonstrate the robustness and generalizability of our overall system over numerous imaging and clinical experiments (\cf Section \ref{sec:ref}).

\section{Methodology}
\label{sec:method}

\noindent\textbf{(1) Multi-modal 2D Keypoint Detector with Attention Fusion}. 
Most recent works in 2D pose estimation \cite{openpose_pami,he2017mask,sun2019hrpose} are essentially single-source (\ie, RGB only) architectures. Consequently, while they work reasonably in generic uncovered patient cases, they fail in more specific ones, \eg, when the patient is covered by a cloth -- a prevalent situation in numerous medical scanning procedures and interventions. 
As existing methods fail to ubiquitously work across imaging modalities and applications, 
we propose a multi-sensory data processing architecture that leverages information from multiple data sources to account for both generic as well as specialized scenarios (\eg, cloth-covered patients). 
We first introduce how to individually train 2D keypoint detectors on single modalities (\eg, RGB or depth), then how to learn complementary information from multiple modalities to improve detection performance and generalizability.

\begin{figure*}[t]
	\centering
 	\includegraphics[width=.9\linewidth]{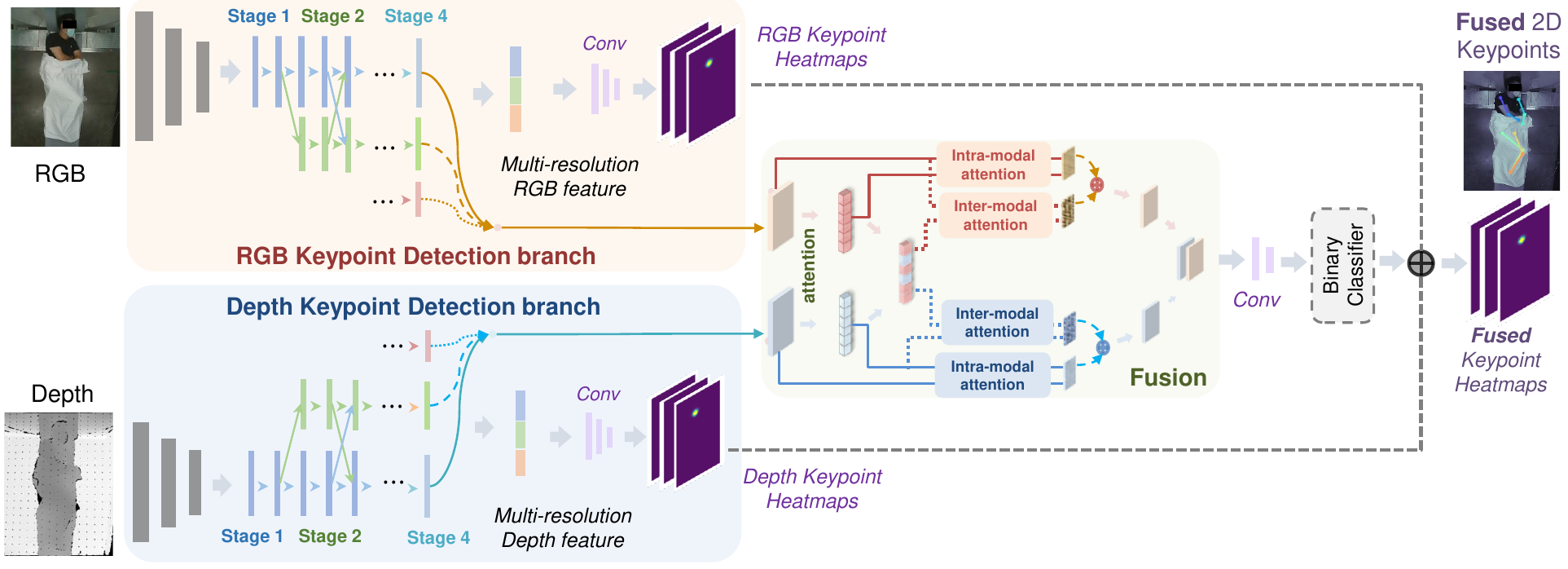}
	\caption{Proposed RGBD keypoint detection framework with attention fusion.}
	\label{fig:fusion}
\end{figure*}

Given an RGB or depth person image, the 2D keypoint detection task aims to predict a set of $N_J$ joint (usually predefined) locations of the person in the image space, which is typically achieved by learning a deep CNN network in most recent works. Here we adopt HRnet \cite{sun2019hrpose} as the backbone architecture which takes the RGB or depth image as input and outputs $N_J$ 2D joint heatmaps, with the peak location of each heatmap $i=1,...,N_J$ indicating the corresponding joint's pixel coordinates, as illustrated in  
Figure \ref{fig:fusion} (orange/blue block for RGB/depth). 

While the training of RGB-based keypoint detector can leverage many publicly available datasets \cite{coco2014,mpii_16cvpr,LSP_Johnson10}, 
the number of depth/RGBD image datasets curated for keypoint detection is much more limited, mainly due to lower sensor accessibility and image quality.
Thus directly training depth-based keypoint detectors over such limited data can easily result in overfitting and poor detection performance during testing. 
To alleviate this, we propose to first utilize an unsupervised pretraining technique \cite{chen2020simsiam} to learn a generalized representation from unlabeled depth images,
which is proved to have better generalizability for downstream tasks like keypoint detection; and then to finetune with labeled training data for improved keypoint detection accuracy. 
This way, we can leverage a larger number of public depth or RGBD image datasets collected for other tasks for a more generic model learning. For further details \wrt the unsupervised pretraining for representation learning, we refer the readers to \cite{chen2020simsiam}. 

Color and depth images contain complementary information of the patient, as well as complementary benefits over different scenarios. \Eg, when the patient is covered by a surgical sheet (Figure \ref{fig:fig1}(b)), RGB features will be heavily affected due to the cover occlusion, whereas depth data still contain rich shape and contour information useful for patient body modeling.
We seek to design an attentive multi-modal fusion network, to effectively aggregate complementary information across RGB and depth images by enforcing intra- and inter-modal attentive feature aggregation for improved keypoint detection performance. Specifically, we propose a two-branch score-based RGBD fusion network as shown in Figure \ref{fig:fusion}. In the proposed fusion network, we take the last stage features of the HRnet backbone from RGB and depth branches respectively, and forward them into a fusion module with intra-modal and inter-modal attentive feature aggregation, for a binary classification score prediction. 
This classifier aims to determine (\cf output score) which modality (RGB or depth) results in the most reliable prediction, based on the prediction error from RGB and depth branches during training. 
For example, if the RGB prediction error is larger than depth prediction error, we set the classifier label to 0, and vice-versa (setting to 1 if RGB-based error is lower).
After the binary classifier is learned, it will produce a probability score (range from 0 to 1) indicating the reliability of RGB and depth branch predictions, which is then utilized to weight keypoint heatmap predictions from each branch before their fusion. 
In this way, the proposed module is able to fuse complementary information from single modalities and learn enhanced feature representations for more accurate and robust keypoint detection. 

\noindent\textbf{(2) Self-supervised 3D Mesh Regressor}. 
After producing the 2D keypoints, we aim to recover the 3D mesh representation of the patient for complete and dense patient modeling. Note that we use the Skinned Multi-Person Linear (SMPL) model \cite{loper2015smpl}, which is a statistical parametric mesh model of the human body, represented by pose $\bm{\theta}\in \mathbb{R}^{72}$ and shape $\bm{\beta} \in \mathbb{R}^{10}$ parameters. Unlike prior works \cite{yang2020robust} that require both images and the corresponding ground-truth 3D mesh parameters for training, our proposed method does not need such expensive annotations and relies only on synthetically generated pairs of 2D keypoint predictions and mesh parameters. Our method thus does not suffer from the biased distribution and limited scale of existing 3D datasets.

Specifically, to generate the synthetic training data, we sample SMPL pose parameters from training sets of public datasets (\ie, AMASS \cite{AMASS:2019}, UP-3D \cite{lassner2017unite} and 3DPW \cite{von2018recovering}), and shape parameters from a Gaussian distribution following \cite{sengupta2020synthetic}.
We then render the 3D mesh given the sampled $\bm{\theta}$ and $\bm{\beta}$ and project the 3D joints determined by the rendered mesh to 2D keypoint locations given randomly sampled camera translation parameters. The $N_J$ 2D keypoint locations then can be formed into $N_J$ heatmaps (as described in Section \ref{sec:method}(1)) and passed to a CNN for $\bm{\theta}$ and $\bm{\beta}$ regression. Here we use a Resnet-18 \cite{He_2016_CVPR} as the baseline architecture for mesh regression. In our experiments, we extensively sampled the data points and generate 330k synthetic data pairs to train the mesh regressor. During testing, the 2D keypoint heatmaps inferred from the RGBD keypoint detection model (\cf Section \ref{sec:method}(1)) are directly utilized for 3D mesh estimation.

\section{Experiments}
\label{sec:ref}
\textbf{Datasets, implementation, and evaluation.} To demonstrate the efficacy of our proposed method, we evaluate on the public SLP dataset \cite{SLP_2019} (same train/test splits from the authors) and proprietary RGBD data collected (with approval  from ethical review board) from various scenarios: computed tomography (CT), molecular imaging (MI), and magnetic resonance imaging (MRI). 

To collect our proprietary MI dataset, 13 volunteers were asked to take different poses while being covered by a surgical sheet with varying covering areas (half body, 3/4 body, full body) and facial occlusion scenarios (with/without facial mask). We use 106 images from 3 subjects to construct the training set, and 960 images from the remaining 10 subjects as test set. 
For the dataset collected in MRI room, we captured 1,670 images with varying scanning protocols (\eg, wrist, ankle, hip, \etc) and patient bed positions, with the volunteers being asked to show a variety of poses while being covered by a cloth with different level of occlusions (similar to MI dataset).
This resulted in 1,410 training images and 260 testing ones. For our proprietary CT dataset, we asked 13 volunteers to lie on a CT scanner bed and exhibit various poses with and without cover. We collected 974 images in total. To test the generalizability of the proposed mesh estimator across imaging modalities, we use this dataset for testing only.

During training stage, we use all data from the SLP, MI and MRI training splits, along with public datasets COCO \cite{coco2014} and MPII \cite{mpii_16cvpr} to learn our single RGB keypoint detector, with the ground-truth keypoint annotations generated manually. For our depth detector, we pretrain its backbone over public RGBD datasets, \ie, 7scene \cite{7scene_cvpr}, PKU \cite{liu2017pku}, CAD \cite{cad_ICRA} and SUN-RGBD \cite{sunrgbd_cvpr}. We then finetune the model over SLP, MI and MRI training data with keypoint supervision. 
We apply the commonly-used 2D mean per joint position error (MPJPE) and percentage of correct keypoints (PCK) \cite{2deval_bm2018} for quantifying the accuracy of 2D keypoints, and the 3D MPJPE, Procrustes analysis (PA) MPJPE \cite{kanazawa2018end} and scale-corrected per-vertex Euclidean error
in a neutral pose (T-pose), \ie, PVE-T-SC \cite{sengupta2020synthetic} (all in mm) for 3D mesh pose and shape evaluation. 

\subsection{2D Keypoint Prediction}
\textbf{(1) Comparison to State-of-the-art}.
In Table \ref{table:test} (first row), we compare the 2D MPJPE of our keypoint prediction module with competing 2D detectors on the SLP dataset. Here, ``Ours (RGB)" and ``Ours (Depth)" refer to the proposed single-modality RGB and depth keypoint detectors, which achieve substantial performance improvement, including compared to the recent RDF algorithm of Yang \etal \cite{yang2020robust}. In Table \ref{table:2dpck_op}, we compare the PCK@0.3 of proposed RGBD keypoint detector with off-the-shelf state-of-the-art 2D keypoint detector OpenPose \cite{openpose_pami} on SLP and MI datasets. We notice that our solution performs significantly better than OpenPose across different data domains, which demonstrates the superiority of the proposed method. We present more PCK@0.3 (torso) evaluations of the proposed multi-modal keypoint detector with attentive fusion in Table \ref{table:2dpck} on MRI and CT (cross-domain) dataset, proving the efficacy of the proposed multi-modal fusion strategy.

\begin{figure}[t]
	\centering
 	\includegraphics[width=.8\linewidth]{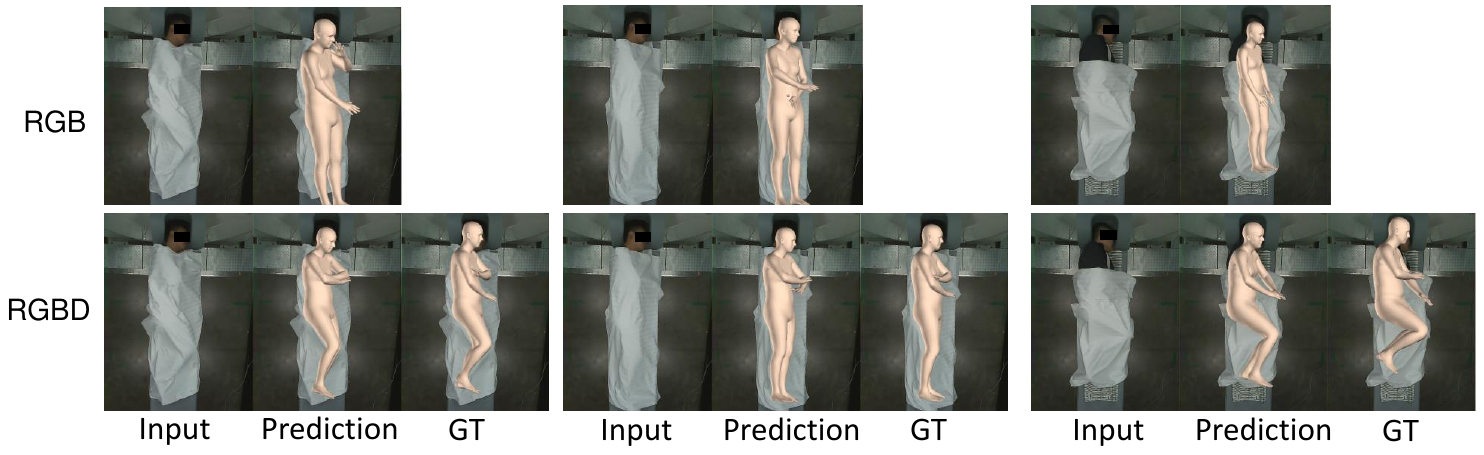}
	\caption{Performance comparison between proposed RGB and RGBD model.} 
	\label{fig:vis4}
\end{figure}

\noindent\textbf{(2) Ablation Study on Multi-Modal Fusion}. Table \ref{table:test2} (A) contains results of an ablation study to evaluate the impact of utilizing single (RGB/depth) and multi-modal fused (RGBD) data for keypoint detection on SLP, MI, MRI and CT (cross-domain) data. We evaluated on different CNN backbones, \ie HRNet \cite{sun2019hrpose} and ResNet-50 \cite{He_2016_CVPR} (see supplementary material), and we observe consistent performance improvement across all datasets with multi-modal fusion, demonstrating the efficacy of our fusion architecture. See Figure~\ref{fig:vis4} for a qualitative illustration of this aspect.

\noindent\textbf{(3) Ablation Study on Unsupervised Pretraining of Depth Keypoint Detector}.
To demonstrate the advantage of utilizing unsupervised pretraining strategy for generalized keypoint detection, another ablation study is performed \wrt our single depth-based detector on SLP and MI data, pretrained with varying number of unannotated data, then finetuned with a fix amount of labeled samples (SLP, MI and MRI data). 
We can see from Table \ref{table:test3} that the keypoint detection performance generally increases along with the quantity of pretraining data, proving the efficacy of the proposed unsupervised pretraining strategy.

\begin{table}[t]
    \centering
    \caption{Comparison on SLP~\cite{SLP_2019} to existing methods, \wrt 2D keypoint detection and 3D mesh regression (modalities: ``RGB" color, ``T" thermal, ``D" depth). Grey cells indicate numbers not reported in the references.}
    \scalebox{0.85}{
    \begin{tabular}{l@{\hskip 5pt}r@{\hskip 5pt}|x|x|xxx|*3{x}|*3{x}}
    \toprule  
    & \multirow{2}{*}{Methods:}
    & \multirow{2}{*}{\makecell[c]{SPIN\\\cite{kolotouros2019spin}}}
    & \multirow{2}{*}{\makecell[c]{OP\\\cite{openpose_pami}}}
    & \multicolumn{3}{c}{HMR \cite{kanazawa2018end}}
    & \multicolumn{3}{|c}{RDF \cite{yang2020robust}}
    & \multicolumn{3}{|c}{\textbf{Ours}}
    \\
    & & & &
    \scriptsize{RGB} & \scriptsize{T} & \scriptsize{RGBT} &
    \scriptsize{RGB} & \scriptsize{T} & \scriptsize{RGBT} &
    \scriptsize{RGB} & \scriptsize{D} & \scriptsize{RGBD} 
    \\  
    \midrule  
    \parbox[t]{2mm}{\multirow{2}{*}{\rotatebox[origin=c]{90}{2D}}}
    &
    MPJPE \footnotesize{(px)}$\downarrow$ 
    &  \cellcolor{gray!25} 
    & 293.8 
    & 37.2 & \cellcolor{gray!25} & \cellcolor{gray!25}
    & 36.6 & \cellcolor{gray!25} & \cellcolor{gray!25} 
    & \textbf{17.1} & \textbf{14.2} & \textbf{13.2} \\
    
    &
    MPJPE \footnotesize{(cm)}$\downarrow$ 
    & \cellcolor{gray!25}
    & 163.9
    & 20.8 & \cellcolor{gray!25} & \cellcolor{gray!25} 
    & 20.4 & \cellcolor{gray!25} & \cellcolor{gray!25}
    & \textbf{9.5} & \textbf{7.9} & \textbf{7.4} \\
    \midrule  
    
    \parbox[t]{2mm}{\multirow{1}{*}{\rotatebox[origin=c]{90}{3D}}}
    & 
    MPJPE \footnotesize{(mm)}$\downarrow$ 
    & 236 
    & \cellcolor{gray!25} 
    & 155 & 149 & 143 
    & 144 & 138 & 137 
    & \textbf{123} & \textbf{118} & \textbf{115} \\
    
    \bottomrule  
    \end{tabular}}
    \label{table:test}
\end{table}

\begin{table}[ht!]
    \centering
    \caption{Ablation study and evaluation on different imaging modalities \wrt 2D keypoint detector (A) and 3D mesh regressor (B).
    \footnotesize{($\dag$ $=$ cross-domain evaluation)}
    }
    \setlength{\tabcolsep}{2pt}
    \scalebox{0.9}{
    \begin{tabular}{c|*3{C{1.4cm}}|*3{C{.9cm}}|*3{C{.9cm}}}
        \multicolumn{1}{c}{} 
        & \multicolumn{3}{c|}{(A) 2D detector ablation study.}
        & \multicolumn{6}{c}{(B) 3D mesh regressor evaluation.}
        \\
        \toprule  
        \multirow{2}{*}{Data}
        & \multicolumn{3}{c|}{2D MPJPE \scriptsize{(px)}$\downarrow$ } 
        & \multicolumn{3}{c|}{3D PA MPJPE \scriptsize{(mm)}$\downarrow$}
        & \multicolumn{3}{c}{3D PVE-T-SC \scriptsize{(mm)}$\downarrow$}
        \\
        &  \scriptsize{RGB} & \scriptsize{D} & \scriptsize{RGBD} 
        &  \scriptsize{RGB} & \scriptsize{D} & \scriptsize{RGBD} 
        &  \scriptsize{RGB} & \scriptsize{D} & \scriptsize{RGBD} 
        \\
        \midrule  
        SLP 
        & 17.1
        & 14.2
        & \textbf{13.2}  
        & 83.4
        & 78.3 
        & \textbf{77.3} 
        & 17.3
        & 14.5 
        & \textbf{13.3} 
        \\
        MI 
        & 13.0
        & 13.6
        & \textbf{12.6
        } 
        & 97.0
        & 101.5
        & \textbf{93.1
        } 
        & 22.9
        & 26.6
        & \textbf{17.7
        }
        \\
        MRI 
        & 7.7
        & 15.6
        & \textbf{7.2
        }
        & 103.1
        & 99.3
        & \textbf{94.3
        } & 19.8
        & 17.8
        & \textbf{15.1
        }
        \\
        \midrule
        CT$^\dag$ 
        & 23.3
        & 22.5
        & \textbf{21.2
        } 
        & 110.9
        & 107.5
        & \textbf{104.3
        } 
        & 17.3
        & 20.2
        & \textbf{17.3
        }
        \\
        \bottomrule  
    \end{tabular}}
    \label{table:test2}
\end{table}

\begin{table}[ht!]
    \centering
    \caption{
    Impact of pretraining data (7scene \cite{7scene_cvpr}, PKU \cite{liu2017pku}, CAD \cite{cad_ICRA},  SUNRGBD \cite{sunrgbd_cvpr}) on MPJPE accuracy (px) of proposed depth-based keypoint detector.
    }
    \setlength{\tabcolsep}{5pt}
    \begin{tabular}{c|*3{c|}c}
    
        \toprule  
        Pretrain Datasets:
        & \cite{7scene_cvpr}
        & \cite{7scene_cvpr}+\cite{liu2017pku}
        & \cite{7scene_cvpr}+\cite{liu2017pku}+\cite{cad_ICRA}
        & \cite{7scene_cvpr}+\cite{liu2017pku}+\cite{cad_ICRA}+\cite{sunrgbd_cvpr}
        \\
        \midrule
        SLP
        & 17.8
        & 14.5
        & 15.1
        & \textbf{13.5}
        \\
        MI
        & 14.5
        & 13.6
        & 13.8
        & \textbf{13.3} \\
        \bottomrule
    \end{tabular}
    \label{table:test3}
\end{table}

\begin{table}[ht!]
    \centering
    \caption{PCK@0.3 evaluation of our proposed 2D keypoint detector with competing methods on SLP (top) and MI (bottom).}
    \scalebox{0.8}{
    \begin{tabular}{c|p{0.85cm}|p{0.85cm}|p{0.85cm}|p{0.85cm}|p{0.85cm}|p{0.85cm}|p{0.85cm}|p{0.85cm}|p{0.85cm}|p{0.85cm}|p{0.85cm}|p{0.85cm}|p{0.85cm}}
    \toprule
    Methods & R.Ak. & R.Kn. & R.H. & L.H. & L.Kn. & L.Ak. & R.Wr. & R.Eb. & R.Sh. & L.Sh. & L.Eb. & L.Wr. & Avg \\
    \midrule
    OpenPose\cite{openpose_pami} & 13.0 & 38.2 & 74.6 & 73.9 & 34.6 & 11.1 & 54.9 & 74.6 & 95.7 & 95.7 & 73.3 & 52.6 & 57.7\\
    \midrule
    \textbf{Proposed} & \textbf{98.4} & \textbf{98.4} & \textbf{100.0} & \textbf{100.0} & \textbf{99.6} & \textbf{98.2} & \textbf{92.5} & \textbf{97.2} & \textbf{99.9} & \textbf{99.3} & \textbf{96.1} & \textbf{94.7} & \textbf{97.9} \\
    \bottomrule
    \end{tabular}}
    \\
    \scalebox{0.8}{
    \begin{tabular}{c|p{0.85cm}|p{0.85cm}|p{0.85cm}|p{0.85cm}|p{0.85cm}|p{0.85cm}|p{0.85cm}|p{0.85cm}|p{0.85cm}|p{0.85cm}|p{0.85cm}|p{0.85cm}|p{0.85cm}}
    \toprule
    Methods & R.Ak. & R.Kn. & R.H. & L.H. & L.Kn. & L.Ak. & R.Wr. & R.Eb. & R.Sh. & L.Sh. & L.Eb. & L.Wr. & Avg \\
    \midrule
    OpenPose\cite{openpose_pami} & 0.0 & 0.0 & 2.7 & 3.3 & 0.0 & 0.0 & 20.0 & 34.4 & 85.3 & 86.7 & 34.5 & 18.1 & 23.7\\
    \midrule
    \textbf{Proposed} & \textbf{97.6} & \textbf{99.3} & \textbf{99.9} & \textbf{99.7} & \textbf{97.2} & \textbf{95.4} & \textbf{91.6} & \textbf{97.8} & \textbf{100.0} & \textbf{99.8} & \textbf{98.7} & \textbf{92.5} & \textbf{97.5} \\
    \bottomrule
    \end{tabular}}
    \label{table:2dpck_op}
\end{table}

\begin{table}[ht!]
    \centering
    \caption{PCK@0.3 (torso) evaluation of our proposed 2D keypoint detector on MRI and CT$^\dag$ (cross-validation: no CT training data used in model learning) testing data.}
    \scalebox{0.8}{
    \begin{tabular}{c|p{0.85cm}|p{0.85cm}|p{0.85cm}|p{0.85cm}|p{0.85cm}|p{0.85cm}|p{0.85cm}|p{0.85cm}|p{0.85cm}|p{0.85cm}|p{0.85cm}|p{0.85cm}|p{0.85cm}}
    \toprule
     & R.Ak. & R.Kn. & R.H. & L.H. & L.Kn. & L.Ak. & R.Wr. & R.Eb. & R.Sh. & L.Sh. & L.Eb. & L.Wr. & Avg \\
     \midrule
    MRI & 97.0 & 98.7 & 99.4 & 99.4 & 98.7 & 98.5 & 96.8 & 98.1 & 99.4 & 99.4 & 98.7 & 96.8 & 98.4 \\
    \midrule
    CT$^\dag$ & 91.3 & 93.2 & 93.9 & 94.0 & 92.5 & 91.3 & 84.9 & 88.6 & 93.6 & 93.9 & 88.9 & 86.2 & 91.0 \\
    \bottomrule
    \end{tabular}}
    \label{table:2dpck}
\end{table}

\subsection{3D Mesh Estimation}
\begin{figure}[ht!]
	\centering
 	\includegraphics[width=.85\linewidth]{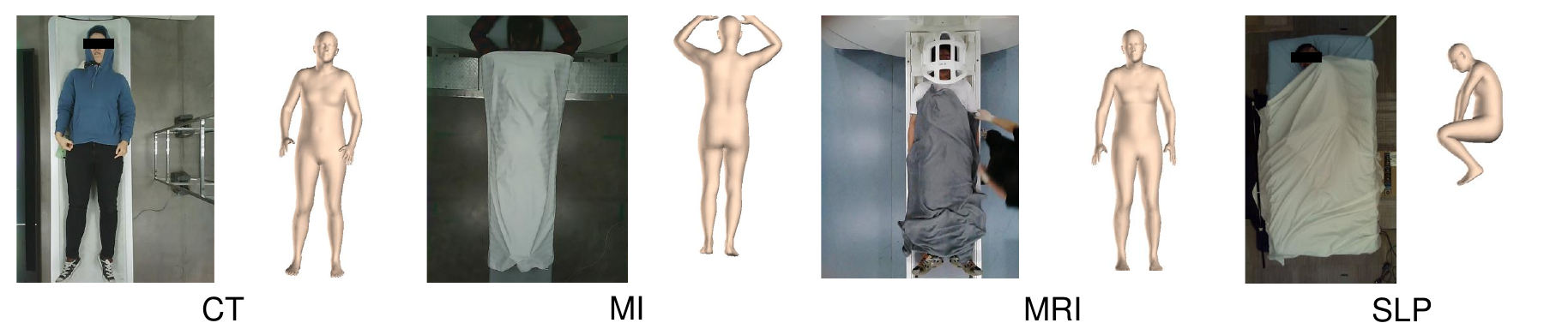}
	\caption{Visualization of reconstructed mesh results on CT, MI, MRI and SLP.}
	\label{fig:vis3}
\end{figure}

We next discuss the performance of our 3D mesh estimation module. To generate ground-truth 3D SMPL pose and shape annotations for all testing data, we apply an off-the-shelf 3D mesh predictor \cite{joo2020eft,SMPL-X:2019} followed by manual refinement. 
Given this testing ground truth, we use the 3D MPJPE and PVE-T-SC metrics to quantify performance. 
Comparison to other 3D mesh estimation technique is shown in Table \ref{table:test} (bottom row). 
Again, we observe substantial performance improvement in terms of 3D joint localization (\cf 3D MPJPE) and per-vertex mesh accuracy (\cf 3D PVE-T-SC) across a wide variety of cover conditions, despite purely relying on synthetic training data (whereas competitive methods require expensive 3D annotations). 
The proposed solution shines (on the SLP data) over the state-of-the-art, \eg, recent method by Yang \etal \cite{yang2020robust} and one of the most commonly used 3D mesh estimator SPIN \cite{kolotouros2019spin}. 
Table~\ref{table:test2} (B) further shows that these improvements are generally consistent across all imaging modalities. Qualitative mesh estimation results are shared in Figure~\ref{fig:vis3}.

\subsection{Automated Isocentering with Clinical CT Scans}
To demonstrate the clinical value of the proposed method, we evaluate the isocentering accuracy in a clinical CT scanning setting. To do so, we mount an RGBD camera above the CT patient support and calibrate it spatially to the CT reference system. With the RGBD images captured by the camera, our proposed method can estimate the 3D patient mesh and compute the thickness of the target body part, which can then be used to adjust the height of the patient support so that the center of target body part aligns with the CT isocenter. We conducted this evaluation with 40 patients and 3 different protocols, and calculated the error based on the resulting CT scout scan as shown in Table \ref{tab:iso}. Compared to the currently deployed automated CT patient positioning system \cite{automatedPP_Euro2021}, our pipeline automatically aligns the center of target body part and scanner isocenter with mean errors of 5.3/7.5/8.1mm for abdomen/thorax/head respectively vs. 13.2mm median error of radiographers in \cite{automatedPP_Euro2021}, which clearly demonstrates the advantage of our proposed positioning workflow.

\begin{table}[ht!]
    \centering
    \caption{Evaluation on ISO-center estimation with clinical CT scans.}
      \scalebox{0.87}{
     \begin{tabular}{c|*5{C{9mm}|}C{9mm}}
     \toprule
    Protocol & \multicolumn{2}{c|}{Abdomen}  & \multicolumn{2}{c|}{Thorax}  & \multicolumn{2}{c}{Head} \\
    \midrule
    \multirow{2}{*}{Error} &Mean & STD & Mean & STD & Mean & STD\\
    \cmidrule{2-7}
    (mm) & 5.3 & 2.1  & 7.5 & 2.9 & 8.1& 2.2 \\
     \bottomrule
     \end{tabular}
     }
      \label{tab:iso}
 \end{table}

\section{Conclusion}
In this work, we considered the problem of 3D patient body modeling and proposed a novel method, consisting of a multi-modal 2D keypoint detection module with attentive fusion and a self-supervised 3D mesh regression module, being applicable to a wide variety of imaging and clinical scenarios. We demonstrated these aspects with extensive experiments on proprietary data collected from multiple scanning modalities as well as public datasets, showing improved performance when compared to existing state-of-the-art algorithms as well as published clinical systems. Our results demonstrated the general-purpose nature of our proposed method, helping take a step towards algorithms that can lead to scalable automated patient modeling and positioning systems.

\bibliographystyle{splncs04}
\bibliography{egbib}

\begin{thebibliography}{10}
\providecommand{\url}[1]{\texttt{#1}}
\providecommand{\urlprefix}{URL }
\providecommand{\doi}[1]{https://doi.org/#1}

\bibitem{2deval_bm2018}
Andriluka, M., Iqbal, U., Insafutdinov, E., Pishchulin, L., Milan, A., Gall,
  J., Schiele, B.: Posetrack: A benchmark for human pose estimation and
  tracking. In: CVPR (2018)

\bibitem{bogo2016keep}
Bogo, F., Kanazawa, A., Lassner, C., Gehler, P., Romero, J., Black, M.J.: Keep
  it smpl: Automatic estimation of 3d human pose and shape from a single image.
  In: ECCV (2016)

\bibitem{automatedPP_Euro2021}
Booij, R., van Straten, M., Wimmer, A., Budde, R.P.: Automated patient
  positioning in ct using a 3d camera for body contour detection: accuracy in
  pediatric patients. European Radiology  \textbf{31},  131–138 (2021)

\bibitem{openpose_pami}
{Cao}, Z., {Hidalgo Martinez}, G., {Simon}, T., {Wei}, S., {Sheikh}, Y.A.:
  Openpose: Realtime multi-person 2d pose estimation using part affinity
  fields. IEEE Transactions on Pattern Analysis and Machine Intelligence
  (2019)

\bibitem{chen2020simsiam}
Chen, X., He, K.: Exploring simple siamese representation learning. CVPR
  (2021)

\bibitem{ching2014patient}
Ching, W., Robinson, J., McEntee, M.: Patient-based radiographic exposure
  factor selection: a systematic review. Journal of medical radiation sciences
  \textbf{61}(3) (2014)

\bibitem{Clever_2020_CVPR}
Clever, H.M., Erickson, Z., Kapusta, A., Turk, G., Liu, K., Kemp, C.C.: Bodies
  at rest: 3d human pose and shape estimation from a pressure image using
  synthetic data. In: CVPR (2020)

\bibitem{survey_2dhpe}
Dang, Q., Yin, J., Wang, B., Zheng, W.: Deep learning based 2d human pose
  estimation: A survey. Tsinghua Science and Technology  \textbf{24}(6) (2019)

\bibitem{georgakis2020hierarchical}
Georgakis, G., Li, R., Karanam, S., Chen, T., Ko{\v{s}}eck{\'a}, J., Wu, Z.:
  Hierarchical kinematic human mesh recovery. In: ECCV (2020)

\bibitem{he2017mask}
He, K., Gkioxari, G., Doll{\'a}r, P., Girshick, R.: Mask r-cnn. In: ICCV (2017)

\bibitem{He_2016_CVPR}
He, K., Zhang, X., Ren, S., Sun, J.: Deep residual learning for image
  recognition. In: CVPR (2016)

\bibitem{LSP_Johnson10}
Johnson, S., Everingham, M.: Clustered pose and nonlinear appearance models for
  human pose estimation. In: BMVC (2010)

\bibitem{joo2020eft}
Joo, H., Neverova, N., Vedaldi, A.: Exemplar fine-tuning for 3d human pose
  fitting towards in-the-wild 3d human pose estimation. In: 3DV (2020)

\bibitem{Kadkhodamohammadi2017_nm}
Kadkhodamohammadi, A., Gangi, A., de~Mathelin, M., Padoy, N.: Articulated
  clinician detection using 3d pictorial structures on rgb-d data. Medical
  Image Analysis  \textbf{35} (2017)

\bibitem{Kadkhodamohammadi2017_tx}
Kadkhodamohammadi, A., Gangi, A., de~Mathelin, M., Padoy, N.: A multi-view
  rgb-d approach for human pose estimation in operating rooms. In: WACV (2017)

\bibitem{kanazawa2018end}
Kanazawa, A., Black, M.J., Jacobs, D.W., Malik, J.: End-to-end recovery of
  human shape and pose. In: CVPR (2018)

\bibitem{karanam2020towards}
Karanam, S., Li, R., Yang, F., Hu, W., Chen, T., Wu, Z.: Towards contactless
  patient positioning. IEEE transactions on medical imaging  \textbf{39}(8)
  (2020)

\bibitem{kolotouros2019learning}
Kolotouros, N., Pavlakos, G., Black, M.J., Daniilidis, K.: Learning to
  reconstruct 3d human pose and shape via model-fitting in the loop. In: ICCV
  (2019)

\bibitem{kolotouros2019spin}
Kolotouros, N., Pavlakos, G., Black, M.J., Daniilidis, K.: Learning to
  reconstruct 3d human pose and shape via model-fitting in the loop. In: ICCV
  (2019)

\bibitem{lassner2017unite}
Lassner, C., Romero, J., Kiefel, M., Bogo, F., Black, M.J., Gehler, P.V.: Unite
  the people: Closing the loop between 3d and 2d human representations. In:
  CVPR (2017)

\bibitem{li2007automatic}
Li, J., Udayasankar, U.K., Toth, T.L., Seamans, J., Small, W.C., Kalra, M.K.:
  Automatic patient centering for mdct: effect on radiation dose. American
  journal of roentgenology  \textbf{188}(2) (2007)

\bibitem{coco2014}
Lin, T., Maire, M., Belongie, S., et~al.: Microsoft coco: Common objects in
  context. In: ECCV (2014)

\bibitem{liu2017pku}
Liu, C., Hu, Y., Li, Y., Song, S., Liu, J.: Pku-mmd: A large scale benchmark
  for continuous multi-modal human action understanding. arXiv:1703.07475
  (2017)

\bibitem{SLP_2019}
Liu, S., Ostadabbas, S.: Seeing under the cover: A physics guided learning
  approach for in-bed pose estimation. In: MICCAI (2019)

\bibitem{loper2014mosh}
Loper, M., Mahmood, N., Black, M.J.: Mosh: Motion and shape capture from sparse
  markers. ACM Transactions on Graphics  \textbf{33}(6) (2014)

\bibitem{loper2015smpl}
Loper, M., Mahmood, N., Romero, J., Pons-Moll, G., Black, M.J.: Smpl: A skinned
  multi-person linear model. ACM transactions on graphics  \textbf{34}(6)
  (2015)

\bibitem{AMASS:2019}
Mahmood, N., Ghorbani, N., F.~Troje, N., Pons-Moll, G., Black, M.J.: Amass:
  Archive of motion capture as surface shapes. In: ICCV (2019)

\bibitem{SMPL-X:2019}
Pavlakos, G., Choutas, V., Ghorbani, N., Bolkart, T., Osman, A.A.A., Tzionas,
  D., Black, M.J.: Expressive body capture: 3d hands, face, and body from a
  single image. In: CVPR (2019)

\bibitem{mpii_16cvpr}
Pishchulin, L., Insafutdinov, E., Tang, S., Andres, B., Andriluka, M., Gehler,
  P., Schiele, B.: Deepcut: Joint subset partition and labeling for multi
  person pose estimation. In: CVPR (June 2016)

\bibitem{sengupta2020synthetic}
Sengupta, A., Budvytis, I., Cipolla, R.: Synthetic training for accurate 3d
  human pose and shape estimation in the wild. In: BMVC (2020)

\bibitem{7scene_cvpr}
Shotton, J., Glocker, B., Zach, C., Izadi, S., Criminisi, A., Fitzgibbon, A.:
  Clustered pose and nonlinear appearance models for human pose estimation. In:
  CVPR (2013)

\bibitem{avatar_Vivek2017}
Singh, V., Ma, K., Tamersoy, B., et~al.: Darwin: Deformable patient avatar
  representation with deep image network. In: MICCAI (2017)

\bibitem{sunrgbd_cvpr}
Song, S., Lichtenberg, S.P., Xiao, J.: Sun rgb-d: A rgb-d scene understanding
  benchmark suite. In: CVPR (2015)

\bibitem{srivastav2018mvor}
Srivastav, V., Issenhuth, T., Kadkhodamohammadi, A., de~Mathelin, M., Gangi,
  A., Padoy, N.: Mvor: A multi-view rgb-d operating room dataset for 2d and 3d
  human pose estimation (2018)

\bibitem{sun2019hrpose}
Sun, K., Xiao, B., Liu, D., Wang, J.: Deep high-resolution representation
  learning for human pose estimation. In: CVPR (2019)

\bibitem{cad_ICRA}
Sung, J., Ponce, C., Selman, B., Saxena, A.: Unstructured human activity
  detection from rgbd images. In: ICRA (2012)

\bibitem{von2018recovering}
Von~Marcard, T., Henschel, R., Black, M.J., Rosenhahn, B., Pons-Moll, G.:
  Recovering accurate 3d human pose in the wild using imus and a moving camera.
  In: ECCV (2018)

\bibitem{yang2020robust}
Yang, F., Li, R., Georgakis, G., Karanam, S., Chen, T., Ling, H., Wu, Z.:
  Robust multi-modal 3d patient body modeling. In: MICCAI (2020)

\bibitem{Multimodal_arxiv2020}
Yin, Y., Robinson, J.P., Fu, Y.: Multimodal in-bed pose and shape estimation
  under the blankets. In: ArXiv:2012.06735 (2020)

\end{thebibliography}

\end{document}


\title{Supplementary Material: Self-supervised 3D Patient Modeling with Multi-modal Attentive Fusion}
%
%
\author{Meng Zheng\inst{1}\orcidID{0000-0002-6677-2017} \and
Benjamin Planche\inst{1}\orcidID{0000-0002-6110-6437} \and
Xuan Gong\inst{2}\orcidID{0000-0001-8303-633X}
\and
Fan Yang\inst{1}\orcidID{0000-0003-1535-447X}
\and
Terrence Chen\inst{1}
\and Ziyan Wu\inst{1}\orcidID{0000-0002-9774-7770}}
\authorrunning{F. Author et al.}
%
\institute{United Imaging Intelligence, Cambridge MA, USA \and University at Buffalo, Buffalo NY, USA\\
\email{\{first.last\}@uii-ai.com, xuangong@buffalo.edu}}
\maketitle              
We perform additional ablation study of proposed 2D Keypoint Detector with ResNet-50 [11] backbone in Table \ref{table:fusion_res}, to further prove the generalization of the proposed multi-modal fusion pipeline.

\begin{table}[ht!]
    \centering
    \caption{
    Ablation study on the proposed multi-modal fusion strategy of proposed 2D keypoint detector with ResNet-50 backbone. Numbers are reported in MPJPE accuracy (px).}
    \setlength{\tabcolsep}{5pt}
    \begin{tabular}{c|*2{c|}c}
        \toprule  
        Data & RGB & D &RGBD
        \\
        \midrule
        SLP
        & 19.3
        & 15.3
        & \textbf{14.7}
        \\
        MI
        & 50.0
        & 19.3
        & \textbf{18.5} \\
        MR
        & 14.8
        & 16.7
        & \textbf{13.0} \\
        \bottomrule
    \end{tabular}
    \label{table:fusion_res}
    \vspace{-.5em}
\end{table}